# Outdoor flocking and formation flight with autonomous aerial robots*

G. Vásárhelyi, Cs. Virágh, G. Somorjai, N. Tarcai, T. Szörényi, T. Nepusz, T. Vicsek

*Abstract*— we present the first decentralized multi-copter flock that performs stable autonomous outdoor flight with up to 10 flying agents. By decentralized and autonomous we mean that all members navigate themselves based on the dynamic information received from other robots in the vicinity. We do not use central data processing or control; instead, all the necessary computations are carried out by miniature on-board computers. The only global information the system exploits is from GPS receivers, while the units use wireless modules to share this positional information with other flock members locally. Collective behavior is based on a decentralized control framework with bio-inspiration from statistical physical modelling of animal swarms. In addition, the model is optimized for stable group flight even in a noisy, windy, delayed and error-prone environment. Using this framework we successfully implemented several fundamental collective flight tasks with up to 10 units: i) we achieved self-propelled flocking in a bounded area with self-organized object avoidance capabilities and ii) performed collective target tracking with stable formation flights (grid, rotating ring, straight line). With realistic numerical simulations we demonstrated that the local broadcast-type communication and the decentralized autonomous control method allows for the scalability of the model for much larger flocks.

## Introduction

Collective motion is one of the most spectacular phenomena in nature, where the local behavior of many autonomous, similar individuals results in complex motion patterns, creating a fluctuating meta-organism [1]. Besides being truly astonishing, flocking has many advantages in natural systems, since a group of animals or people is: i) more effective when solving foraging or navigational tasks [2]; ii) robust in terms of the knowledge the flock holds about its actual goal due to redundancy and the system's intrinsic parallel-processing structure; iii) more alert regarding environmental threats and more defensive against attacks [3] [4]. Such natural systems provide bio-inspiration for swarm robotics aiming at creating nature-like, i.e., efficient flocks of artificial units with decentralized control.

There are already a great number of two-dimensional ground-based robotic swarms, being able to perform intriguingly complex tasks, such as playing soccer [5], carrying objects together [6], being self-organized [7], performing self-assembly [8] or simply being large-scale [9] [10]. Implementing robots capable of moving in three-dimensions has a great extra potential. However, achievements concerning aerial flocks have been rather limited so far. This fact comes partly from the current technological difficulties of creating even one reliable flying object that is capable of autonomously performing all the necessary flight maneuvers, such as stable hovering, flying at predetermined speed in a given direction or adapting to a time-dependent altitude and/or flight direction quickly enough. However, the recent appearance of small-scale, lightweight flying robots, commonly named as multi-copters, seems to be filling this niche perfectly. Another severe technical difficulty is the proper perception of the environment and other flock members, i.e., sensing absolute or relative position reliably. All current solutions to this challenge are limited and depend on state-of-the-art technology: computer-vision, infrared, ultrasonic, laser or digital wireless. In order to overcome some of the constraints of these approaches, we have chosen a combination of GPS devices and wireless communication to obtain local relative positional data. In this way we could focus on our primary goal, the implementation of our control algorithms into a real, functional outdoor system.

There have been several remarkable attempts for the creation of 3D flying flocks. Welsby et al. used three motorized balloon-like objects that were wandering in the air together indoor [11]. De Nardi and Holland proposed (but did not create) a 3D flock of helicopters [12]. Hauert et al. presented an autonomous flock of 10 fixed-wing UAVs flying at relatively large distance from each other [13]. Their system is a noteworthy example of an autonomous Reynolds-flock [14], although their UAVs did not have the eventual constraint of avoiding each other since they flew at different fixed altitudes, i.e., repulsion and mutual distance between units was not optimized to prevent crashes. The same applies for [15] where 3 UAVs are said to be flocking in a distributed sensing task but there were no actual interactions between units, they simply flew independently at the same time and same place, at different altitudes. As we will see, repulsive interaction between units that is strong enough to hinder collisions in any critical situation is the origin of self-excited oscillations, which are quite challenging to handle, if not the most difficult task in a delayed environment.

Stirling et al. stated that they created an autonomous indoor quadrotor flock of 3 units. In fact, two of the robots were attached to a ferromagnetic ceiling and only one unit was flying at a time, using the other two as reference points for navigation [16]. However, their relative positioning capability is noteworthy. Kushleyev et al. (V. Kumar Lab) created the most remarkable flock of 20 miniature

*Research supported by the EU ERC COLLMOT project.
G. V. Author was partly supported by EU TÁMOP 4.2.4.A/1-11-1-2012-0001.

All authors are with the Department of Biological Physics, Eötvös University, Budapest, Hungary (corresponding e-mail: vasarhelyi@angel.elte.hu)
G. V., G. S., T. N. and T. V. Authors are with the MTA-ELTE Research Group for Statistical and Biological Physics, Budapest, Hungary

quadcopters, but their system is not autonomous by our definition since a central computer calculates the navigational instructions for all robots [17]. Another limitation of the general applicability of their system is that a fixed, indoor, high-precision positioning system (VICON) is required for any flight. . Bürkle et al. created an outdoor quadcopter swarm, but again, with intensive central processing at a ground station [18].

The implementation that is probably closest to our approach is from Hoffmann et al. (the STARMAC project), in which an autonomous flock of three quadrotors was shown that could perform collision-free flight using algorithms based on Nash Bargaining [19].

The theory of flight formation control, especially of UAVs is widespread [20] [21]. However, experimental results on autonomous unmanned vehicles in formation are very limited. One nice example is given in Turpin et al. (V. Kumar Lab) [22] where four quadrotors are flying in a rectangular or linear shape, indoor, using a VICON system for position reference. Their decentralized control algorithm is actually executed on a central computer, but the results are promising.

Table I summarizes all mentioned state-of-the-art approaches for 3D multi-vehicle implementations.

In this paper we present the first outdoor 'GPS-vision' based swarm of ten autonomous flying robots with decentralized hardware and self-control and stable self-organization capabilities for flocking, target tracking and formation flights. The most important advantages of our approach compared to the state-of-the-art are the following:

- there is true local interaction between units for self-organized collision-avoidance
- we control 10 units instead of 3-4, which is challenging both theoretically and technically
- our solution works outdoor anywhere on the planet where GNSS satellites are available
- our control as well as the system itself is decentralized, i.e. no central computer is needed for operation
- finally and most importantly: our solution is optimized for relatively large inner and outer errors (noise, delay, limited communication range)

Our approach was to use commercially available, affordable, open source products as base units, already having a low-level control of flight stability. We have expanded these units with our self-developed autopilot board and custom-designed flocking algorithms for the high-level control of the units within the swarm.

STRUCTURE OF A UNIT

*A. Basic Unit*

Our flying robots are based on a partially open source quadcopter, called MK Basicset L4-ME, from MikroKopter Co., Germany [23] (Fig. 1), capable of performing onboard self-stabilization and altitude hold. In its original form, it is controlled manually with a standard R/C remote controller.

The main board of the copter accepts virtual controller signals (pitch/roll/yaw and altitude setpoint) from an arbitrary extension board; we used this channel to gain full automatic control over the unit, while maintaining the option of manual interruption with the remote controller.

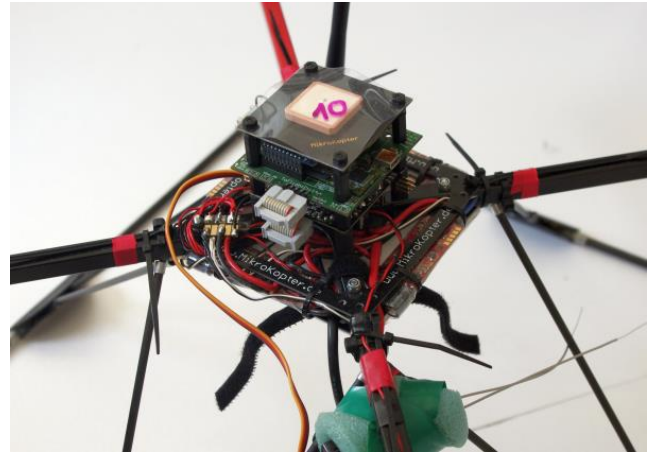

Figure 1. Image of a unit with the commercial base copter below, the custom made (green) extension board above and the GPS on top. Take-off mass is around 1 kg, tip-to-tip diameter is 80 cm. We achieved a maximum of 20 minutes flying time by using 3000mAh, four-cell LiPo batteries.

TABLE I. COMPARISON OF 3D MULTI-VEHICLE IMPLEMENTATIONS

| Authors | Year | Vehicle | N | Decentra-lized? | Collision-avoidance? | Terrain | Dependency | Uniqueness |
|---|---|---|---|---|---|---|---|---|
| Welsby et al. | 2001 | helium balloon | 3 | yes | no | indoor | arena | first 3D, relative IR positioning |
| Hauert et al. | 2011 | fixed-wing | 10 | yes | weak / not crucial | outdoor | GPS | first 10 autonomous |
| Bürkle et al. | 2011 | quadrotor | 5 | no interactions | no | outdoor | GPS, ground control station | extendible framework |
| Hoffmann et al. | 2011 | quadrotor | 3 | yes | yes | indoor / outdoor | GPS+base station outdoor, over-head camera indoor | extendible framework, nice vehicle dynamics |
| Kushleyev et al. | 2012 | quadrotor | 20 | no | n.a. | indoor | VICON, central computer | 20 units, best precision control |
| Turpin et al. | 2012 | quadrotor | 4 | SW yes HW no | yes | indoor | VICON, central computer | quick dynamics |
| Stirling et al. | 2012 | quadrotor | 3 | yes | n.a. | indoor | ferromagnetic ceiling | relative positioning |
| Quintero et al. | 2013 | fixed-wing | 3 | no | no | outdoor | GPS, ground control station | distributed sensing |
| Vasarhelyi et al. (this paper) | 2014 | quadrotor | 10 | yes | yes | outdoor | GPS | see list in the text |

## B. Processor Board

We have developed an extension board (called FlockControl) that contains a 3D gyroscope, a 3D accelerometer, a 3D magnetometer, a pressure sensor, a GPS receiver, a 2.4GHz XBee unit (for digital wireless communication) and a GumStix Overo Water mini-computer with standard Linux OS. The extension board connects to the main board through SPI. It receives information from the base unit about error conditions and inner state variables (attitude, heading, battery level, remote controller signals) and communicates with other flock members via XBee (sharing ID, position, velocity, attitude and status info). The extension board feeds these inputs into the flocking algorithm and sends the calculated virtual steering signals to the low level control board. The main refresh rate of the steering signals is 40 Hz (Fig. 2), individual GPS refresh rate is 5 Hz and XBee communication rate is 10 Hz.

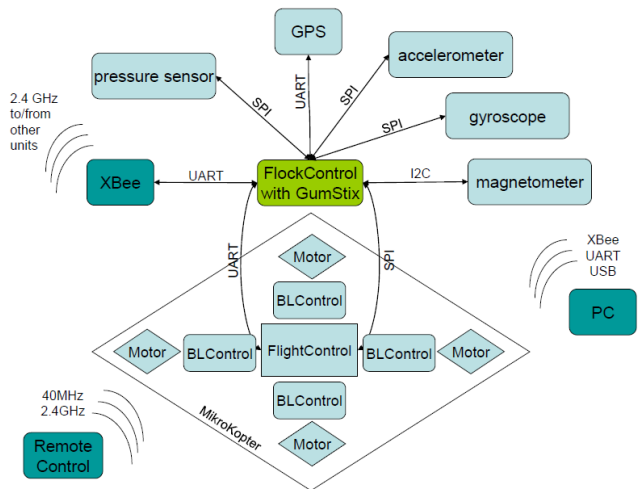

Figure 2. The main structure of one unit. Communication methods between parts are indicated on the arrows.

## C. Communication

The XBee modules work in broadcast sending mode, units send data without establishing one-to-one connection or mesh network with other units. Flock members process incoming XBee packets only from other robots, which are inside their communication range (typically around 50–100 m). That is similar to the way e.g. birds fly in a flock, mostly using the information from neighboring birds [24]. This locality of the communication is a simple way to keep the network scalable to larger flock sizes.

With ten active units the typical delay of the communication (in the process of GPS reception onboard → XBee broadcast → XBee reception on another unit) in our current system is around 0.4s ± 0.2s. However, outages even in the range of seconds can also occur at random times due to the fact that the information is processed by a CPU running Linux with its own priority list of tasks. We plan to shorten delays by using further optimized codes and priority lists. The ultimate solution will be the processing of the signals in a real time operating system (needs new hardware).

## D. Velocity Control

Automatic steering signals are calculated in a world reference frame. The three axes of the control (North, East, Down) are independent from each other, due to the structure of the quadcopter. In the vertical direction we rely on the original altitude-hold algorithm of the L4-ME copter that maintains a pre-defined altitude during most of the flight. Autonomous self-organization thus takes place in a quasi-two-dimensional space, which is satisfactory for many applications, like area coverage or distributed search. To take advantage of the third dimension, during the development phases we defined a slightly different altitude set point for all copters (in the 5–15 m range above ground) to have a safety height distance between units, in case unexpected horizontal control errors would occur. However, we managed to carry out the latest flights (presented in the results section) with all units at the same altitude, with true repulsion and collision-avoidance between units in the horizontal plane.

For the two horizontal axes we implemented a standard PID based velocity control algorithm that converts the target velocity output of the flocking algorithm into the virtual steering signals. Two independent PID controllers are used for the two global axes (North and East) and the target steering signal is converted into a body-fixed coordinate system based on the heading information calculated from the signals of the magnetometer and the other inertial sensors.

Since the $P$ and $D$ terms of the velocity PID tend to zero as the measured velocity is getting closer to the target velocity, the $I$ term must maintain the necessary output signal level for constant speed flight. When flying in a tight flock, quick response is essential; therefore, we have added a linear feed-forward (also known as bias) term to the control loop, by which the system can predict the magnitude of the necessary adjustment based on earlier experience. This way, the $I$ term only needs to fine-tune the estimated stick response and thus the relaxation time of the control loop is significantly reduced. To minimize oscillations of an individual copter, we tuned the PID for slower responsiveness and avoided overshoot of the control.

The performance of the velocity controller was tested in a mobile target tracking setup and it was found that the system is stable and performs well even in moderate wind (up to 5 m/s) with an approximately 1.5 s overall time lag. It is important to note that the PID loop is responsible only for a minor portion of this lag. Other delay-inducing factors are the reaction time of the base unit to the steering signal, the low GPS update rate and the limited acceleration due to the inertia of the copters (Fig. 3).

A price for the stable and overshoot-free, but slower control is paid back at the flock's level, where interesting chaotic oscillations can emerge due to delays, similarly to ghost traffic jams on highways, caused by the slow reaction time (~1s) of humans [25]. These oscillations had to be handled by optimizing our flocking algorithm.

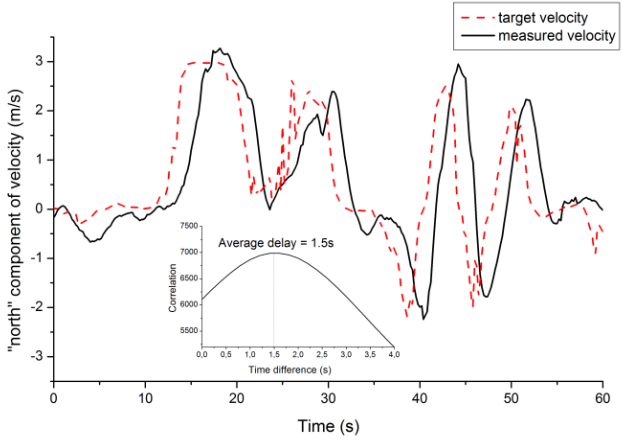

Figure 3. Snapshot of a general target tracking measurement with PID velocity control (*P*=30, *D*=30, *I*=2). Main figure shows the measurement in time, the inset shows the correlation between target velocity (the input of the PID controller) and measured velocity (the final output of the system) as a function of time difference between them. An average time lag of around 1.5 s is present in the control system even if tuned well, due to PID relaxation time, base unit reaction time and inertia limited acceleration.

## OTHER SYSTEM COMPONENTS

### Ground Control

We have also developed a ground station console application for real-time monitoring and debugging purposes. The application runs on an external computer and gathers general telemetry information from all flock members through XBee. Besides monitoring the state of the flock, the ground station can be used for several other purposes, like requesting debugging packets from individual copters with detailed information, changing system parameters in flight to tune behavior, batch uploading new settings, simulate errors or force landing of individuals or all units at once, too. These are all essential features when one works with a large group of robots at once.

### Simulation Framework

We have developed a realistic simulation framework to test various flocking algorithms and to optimize their parameter space before flying with real robots. One main goal of the simulations was to investigate the effects of all sorts of delays present in the system, i.e., reaction time arising from communication and data processing and relaxation time of the controller. A flocking algorithm had to be found that is not sensitive to large delays in terms of stability, in other words, in which delays do not generate undesired oscillations. This is a non-trivial problem, since most bio-inspired flocking algorithms are based on the assumption that reaction times of animals is very small or negligible, i.e., oscillations do not appear at all. To develop robust and oscillation-free flocking algorithms, we combined several techniques, such as using smooth functions instead of sharp ones, using slack in potential valleys or using special over-damped dynamics. These seemingly minor enhancements are part of the novelty of our approach and are actually essential for creating a stable, functional flock of flying robots.

The simulation framework also takes into account many of the inaccuracies of a real system: the position and velocity measurement error, general Gaussian noise corresponding to the environmental effects, the low update rate of the GPS and the limited range of the communication. The final algorithms had to remain stable in that error-prone, realistic framework, as well. More details about the simulation framework and the stability optimization of the algorithms are discussed in a separate paper [26].

## DECENTRALIZED CONTROL ALGORITHM

The control algorithm has a structure analogous to that of the first flocking model of Reynolds [14]: a self-propelled particle needs short-range repulsion from other units, medium-range velocity alignment with neighbors and a global positional constraint to remain with the flock. However, finding the best candidate terms for these tasks in a real environment is not trivial and could be application dependent.

Below we specify how the algorithm described in [26] was implemented for the real flocking and formation flights in the present work. The basic model we use consists of the following main parts (upper indices in the equations mark individuals in the flock).

### Short-range Repulsion: Pair Potentials

A repulsive distance-based potential acts between all close units to avoid collisions:

$$\vec{a}_{pot}^i = \begin{cases} -D \sum_{j \neq i} min(r_1, \ r_0 - |\vec{x}^{ij}|) \frac{\vec{x}^{ij}}{|\vec{x}^{ij}|}, \\ \qquad \text{if } |\vec{x}^{ij}| < r_0, \\ 0 \text{ otherwise,} \end{cases} \quad (1)$$

where $D$ is the spring constant of a repulsive half-spring, $\vec{x}^{ij} = \vec{x}^j - \vec{x}^i$ is the difference of positions of unit *i and j*, $r_0$ is the equilibrium distance, i.e., the distance above which there is no repulsion between units and $r_1$ is used to define an upper threshold for the repulsion to avoid over-excitation of units.

In time-lagged systems with even small reaction time, potential functions in the control can be the source of self-excited oscillations [27]. Even though we used only linear and limited repulsive terms, oscillations emerged in some situations. To avoid these, we forced over-damped dynamics through the next term, the velocity alignment.

### Middle-range Velocity Alignment: Viscous Friction

Units close to each other damp their velocity *difference* to reduce oscillations and to synchronize collective motion with a viscous friction-like term, similar to the one in [28] or [29]:

$$\vec{a}_{slip}^i = C_{frict} \sum_{j \neq i} \frac{\vec{v}^{ij}}{\left(max(|\vec{x}^{ij}| - (r_0 - r_2), \ r_1)\right)^2}, \quad (2)$$

where $C_{frict}$ is the viscous friction coefficient, $\vec{v}^{ij} = \vec{v}^j - \vec{v}^i$ is the velocity difference between units *i* and *j*, $r_2$ defines a constant slope around the equilibrium distance $r_0$, finally, $r_1$ defines a threshold again to avoid division by close-to-zero distances due to e.g. measured GPS position error.

Note that equations 1 and 2 are both dependent of $r_0$ to allow for the dynamic tuning of flock density. This is essential when e.g. one needs to extend the model to higher velocity ranges, where larger 'breaking distance' is needed between units.

*Global Positional Constraint I: Flocking*

One way of keeping individual robots together is to define a bounded area around a global reference point and implement a general self-propelled flocking model [24] that makes the units move around within the walls of the area. The self-propelling term is defined as

$$\vec{v}_{spp}^i = v_{flock} \frac{\vec{v}^i}{|\vec{v}^i|}, \quad (3)$$

where $v_{flock}$ is a constant flocking speed the units try to maintain. The bounding walls are pre-defined globally, but they appear as local attractive shill agents [30] that try to pull units back towards the center of the flight area through virtual velocity alignment:

$$\vec{a}_{wall}^i = C_{shill} f(|\vec{x}_{trg} - \vec{x}^i|, R, d) \left( v_{flock} \frac{\vec{x}_{trg} - \vec{x}^i}{|\vec{x}_{trg} - \vec{x}^i|} - \vec{v}^i \right), (4)$$

where $C_{shill}$ is the viscous friction coefficient of the wall, $\vec{x}_{trg}$ is the center of the flight area, i.e., the position of a global reference/target point and $f(x)$ is a smooth transfer function:

$$f(x, R, d) = \begin{cases} 0 & if\ x \in [0, R] \\ \frac{\sin(\frac{\pi}{d}(x-R)-\frac{\pi}{2})+1}{2} & if\ x \in [R, R+d] \\ 1 & if\ x \in [R+d, \infty] \end{cases} \quad (5)$$

where $R$ is set to be the distance between the reference point and the wall and $d$ defines the width of the decay, i.e., the softness of the wall.

With this definition the soft-wall interacts with the units in a smooth way, only when they are outside of the flight area. Object avoidance in general can also be introduced in a similar way with *repulsive* shill agents.

Note that in the above model all interaction terms are local or quickly decay with distance, corresponding to the short range of the communication. Scalability to larger flock sizes is analyzed in realistic numerical simulations.

*Global Positional Constraint II: Formation Flights*

Another option for global attraction can be introduced without explicit self-propelling, through formation flights around a (static or dynamic) global reference/target point. For this, we split target tracking into two parts:

$$\vec{v}_{track}^i = \begin{cases} v_0 \frac{\vec{v}_{shp}^i + \vec{v}_{trg}^i}{|\vec{v}_{shp}^i + \vec{v}_{trg}^i|} & if\ |\vec{v}_{shp}^i + \vec{v}_{trg}^i| > v_0 \\ \vec{v}_{shp}^i + \vec{v}_{trg}^i & otherwise \end{cases}, \quad (6)$$

where $\vec{v}_{shp}^i$ defines how units arrange themselves into shapes relative to the individually calculated center of mass:

$$\vec{v}_{shp}^i = \beta v_0 f(|\vec{x}_{shp}^i - \vec{x}^i|, R, d) \frac{\vec{x}_{shp}^i - \vec{x}^i}{|\vec{x}_{shp}^i - \vec{x}^i|}, \quad (7)$$

while $\vec{v}_{trg}^i$ defines how the center of mass follows the target with smoothly adjusted, variable speed:

$$\vec{v}_{trg}^i = \alpha v_0 f(|\vec{x}_{trg} - \vec{x}_{COM}^i|, R, d) \frac{\vec{x}_{trg} - \vec{x}_{COM}^i}{|\vec{x}_{trg} - \vec{x}_{COM}^i|}. \quad (8)$$

Constants $\alpha$ and $\beta$ in the range of [0,1] define the strength of the velocity components relative to $v_0$, a maximal tracking velocity, $\vec{x}_{COM}^i$ is the locally calculated center of mass, $\vec{x}_{shp}^i$ is the desired position in the formation around $\vec{x}_{COM}^i$ and $f(x,R,d)$ is the smooth transfer function introduced in equation 5 to avoid abrupt changes in the dynamics.

For basic shapes, like a straight line, a (rotating) circle or a simple lattice-like grid, one piece of global information is needed: N, the number of robots, which defines the size of the formation. The equations defining $\vec{v}_{shp}^i$ are the following:

i) for a simple *grid*, we have $\vec{x}_{shp}^i = \vec{x}_{COM}^i$ and $R = \frac{r_0}{2} g(N) - \frac{r_0}{2}$, where $g(N)$ is a heuristic function defining the radius of the smallest circle that can contain N unit circles [31]. This way all units can fit in the target area around the center of mass of the flock tightly packed. The grid-like arrangement is settled due to the pairwise repulsion and the velocity alignment.

ii) for a *ring*, we define $\vec{x}_{shp}^i$ as the intersection of the ring around $\vec{x}_{COM}^i$ and the bisectrix of the two closest neighbors defined with a simple angle-based signed metric around the center of mass. The radius of the ring has to be defined to fit N robots on its contour. $R$ is set to a small value or zero as the transfer function is used for individual position tracking now. A rotating ring can be achieved with a simple self-propelling term in the tangential direction relative to $\vec{x}_{COM}^i$. The direction of rotation can be chosen in a self-organized manner according to the direction of the average tangential velocity of all units.

iii) for a *line* we define the signed neighborhood metric as the projected distance on the line from $\vec{x}_{COM}^i$ and set $\vec{x}_{shp}^i$ as the average position of the two neighbors on the line. Units with only one or zero neighbors must approach the closest end of the line to ensure that units settle with the equilibrium distance ($r_0$) between them. $R$ is set to a small value or zero again. The angle of the line can be defined in a self-organized way from a linear fit of actual positions.

*Full Dynamic Equation*

The final form of the velocity evolution is similar to the so-called optimal velocity model describing traffic jams [25] or escape-panic [32]:

$$\vec{v}^i(t + \Delta t) = \vec{v}^i(t) + \frac{1}{\tau}\left(\vec{v}_{spp}^i + \vec{v}_{track}^i - \vec{v}^i(t)\right)\Delta t + \left(\vec{a}_{pot}^i + \vec{a}_{slip}^i + \vec{a}_{wall}^i\right)\Delta t, \quad (9)$$

where $\tau$ is the characteristic time needed to reach the velocity $\vec{v}_{spp}^i + \vec{v}_{track}^i$.

Equation 9 is used as the main differential equation in the numeric simulation and also as the main function determining the desired velocity of a copter. Note that in equation 9 the

SPP flocking and formation-flight models are mixed, but they are usually used separately, depending on the actual task of the flock. Separation is achieved by setting $v_{\text{flock}}$ and $v_0$ to the appropriate value.

To compensate for the time-lag previously described, the PID controller input $\vec{v}^i(t + \Delta t)$ is always calculated for a time instance far ahead $t$. In other words we tell the system in the present what to prepare for in the future, allowing it to react in time. In the simplest linear approximation this is achieved by selecting a large $\Delta t$ (in the magnitude of 1 s or so).

Note that the final target velocity is always restricted in the simulations as well as in the real flocking algorithm. This limits the speed of the units for safety reasons.

Table II summarizes optimal parameter choice for the algorithms.

## RESULTS

We successfully established the first decentralized, autonomous multi-copter flock in an outdoor environment, with swarms of up to 10 flying robots (Fig. 4), flying stably for up to 20 minutes.

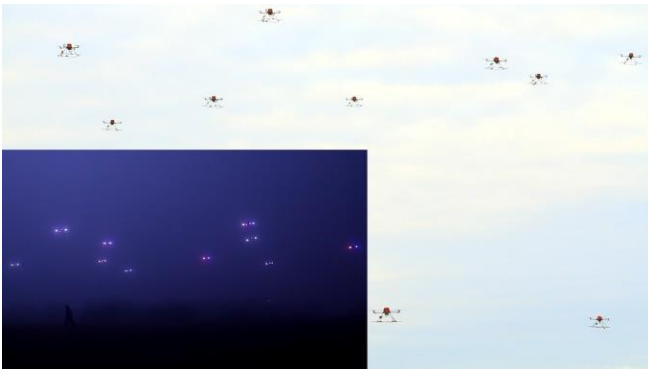

Figure 4. Snapshots of quadcopter flocks of 10 units. Nightlight image: robots are following a target in a grid formation ($r_0$ = 6 m, $v_0$ = 3 m/s). Daylight image: SPP model ($r_0$ = 10 m, $v_{\text{flock}}$ = 2 m/s).

We implemented a flock of two hierarchical levels with a leader robot moving along a pre-defined rectangular path and 9 other robots following the leader in a stable, tight grid formation (Fig 5). To assess the coherence of the flock, we calculated the velocity correlation of the robots as a simple order parameter in the range of [-1, 1]:

$$\Phi = \frac{2}{N(N-1)} \sum_{i<j} \frac{\vec{v}^i \vec{v}^j}{|\vec{v}^i||\vec{v}^j|}. \quad (10)$$

The resulting high velocity correlation indicates the stability of the flock (see caption of Fig 5. for details).

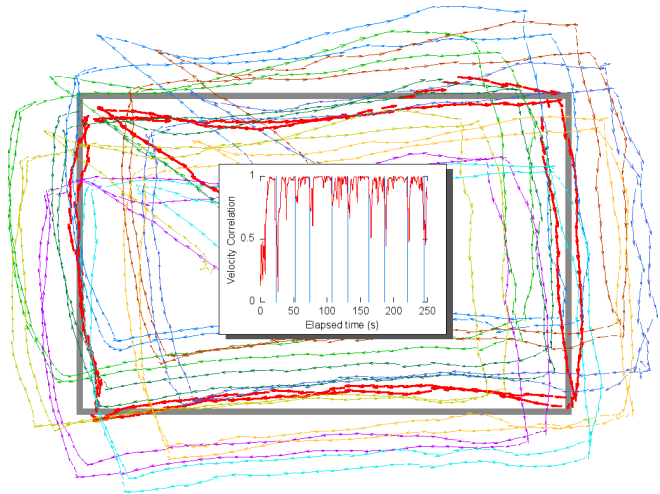

Figure 5. GPS tracklogs of 9 quadcopters in a flock flight, following a 10$^{\text{th}}$ leader kopter as a target (red thick arrows). The target takes off from the center and follows a pre-defined rectangular path (grey rectangle) twice. Inset shows the velocity correlation ($\Phi$) of the flock, with $\overline{\Phi} = 0.84 \pm 0.21$ for the t=[25, 250]s period (without take-off). The low-correlation spikes correspond to the direction change at the corners of the path (indicated by blue lines in the inset). The smooth tracklogs and the high velocity correlation are both indications of stable flocking. Average wind speed was less than 3 m/s. Waypoint rectangle size is 60x40 m. Other parameters: $v_0$ = 2 m/s; $r_0$=10 m.

We successfully implemented the same setup with formation flights of rotating ring and line shapes, as well (Fig. 6). We could change equilibrium distance and switch between shapes in real-time via the R/C remote controller to achieve true self-organization of the formations while adapting to the new situations. General stability is quantified by the standard deviation of the closest neighbor distances of all units, which can be as small as 1 m throughout 1-10 minute periods of a maintained formation (Fig. 6).

TABLE II. OPTIMAL VALUES OF PARAMETERS

| Name | Value | Unit | Comment |
|---|---|---|---|
| pair potential lattice constant ($r_0$) | 6-12 | m | Slack is given due to the large GPS position estimation error |
| pair potential spring constant ($D$) | 1 | 1/s$^2$ | |
| friction coefficient ($C$) | 10–20 | m$^2$/s | Higher values result in more over-damped, slower but more stable flocks |
| preferred velocity ($v_{\text{flock}}$, $v_0$) | 2–4 | m/s | Note that the max flocking/tracking velocity should be smaller than the max allowed velocity |
| TRG coefficient ($\alpha$) | 1 | | A slightly different behavior is achieved if $\alpha + \beta = 1$ and there is no need for cut-off |
| COM coefficient ($\beta$) | 1 | | A slightly different behavior is achieved if $\alpha + \beta = 1$ and there is no need for cut-off |
| transfer function deceleration slope ($v_0/d$) | 0.4 | 1/s | note that this fixed slope is defined on the distance-velocity plot |

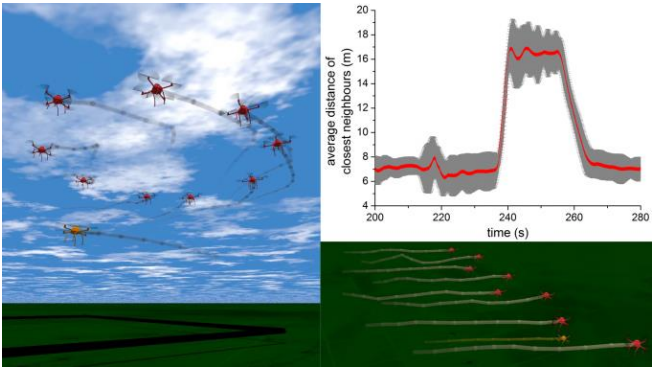

Figure 6. 3D visualization snapshots based on GPS tracklogs of a real target tracking setup with formations of a rotating ring (left) and a line (bottom right). The yellow copter is the leader moving above a pre-defined path (black line), the red ones follow the leader with stable 2D formations. Top right: Average (red curve) and standard deviation (grey fields) of closest neighbor distances in the rotating ring setup. Equilibrium distance is increased from 7m to 17m and back through the R/C remote controller. The small standard deviation (comparable to GPS position error) indicates the stability and smoothness of the formation. $v_{rotation} = 2$ m/s.

The last setup we tested was a self-propelled model inside a ring shaped area ($r_{in}$=15 m, $r_{out}$=45 m). Fig. 7 shows consecutive periods of an emerging self-organized flock of 9 copters circling inside the flight area just like boats [33] or locusts [34] do in a similar setup.

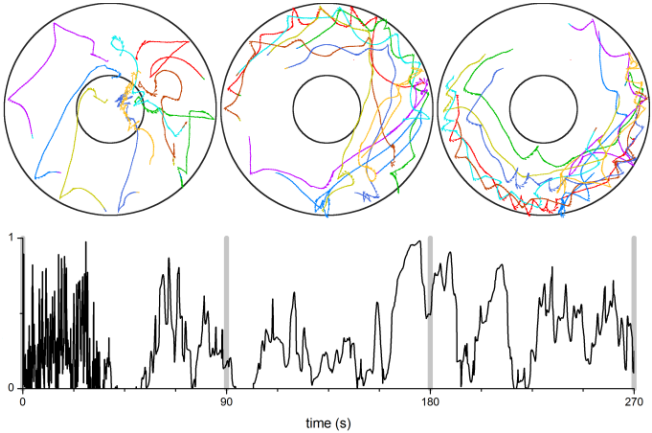

Figure 7. Top: GPS tracklogs from three consecutive 90 s periods of a SPP flock in a circular area after take-off. Bottom: velocity correlation corresponding to the three trajectory plots. Correlation cannot get high for long periods due to the circular area constraint, but the flock tends to organize itself into a coherent swirling motion inside the arena. $r_0$=10 m; $v_0$ = 2 m/s.

Finally, we performed numerical simulations to investigate the robustness of the system, especially when the characteristic size of the expected flock is larger than the communication range $r_C$. Stability of flocking behavior can be measured locally with the parameter below:

$$\Phi(r) = \frac{1}{T}\int_0^T \langle \frac{\vec{v}^i \vec{v}^j}{|\vec{v}^i||\vec{v}^j|}\rangle_{|\vec{r}^{ij}|<r}. \qquad (11)$$

$\Phi(r) \approx 1$ means that the velocity vectors of units close to each other (closer than $r$) are parallel during the simulated experiment. According to Fig. 8, stable flocks can be observed with characteristic size greater than $5r_0$, if $r_C > 3r_0$.

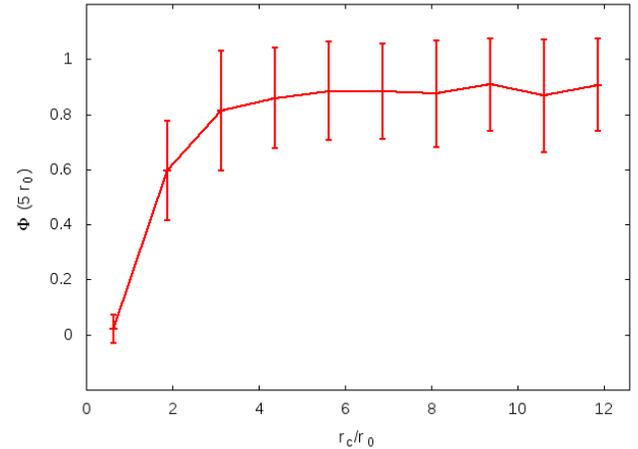

Figure 8. Simulation results about $\Phi(r = 5r_0)$ as a function of communication range $r_C$. All data points are averaged over a 10 minute simulated experiment with 100 agents, within a 300 m wide square-shaped area and random initial conditions. Standard deviation is given as error bars. The environmental effects are modeled with a Gaussian noise term added to the dynamics with $\sigma = 0.1 m^2/s^2$. The time delay of the communication is 1s. Other parameters: $C_{frict} = 30 \frac{m^2}{s}, r_0 = 8m, D = 1s^{-2}, v_{flock} = 2\frac{m}{s}, \Delta t = 1s, \tau = 1s$.

## CONCLUSIONS

Decentralization in multi-robotic systems is one promising way of the future. Accordingly, our flocking algorithms run locally on every robot and do not rely on a ground station or any central data processing or control. We demonstrated the usability of our system with widely available GPS-based positioning; however, the framework is general, the algorithms work with any other sensory input from which relative position, velocity and attitude information can be obtained. Moreover, our approach results in a general high-level hardware and software control layer that can be easily transported to other swarming systems, as well.

Stability of the flock fundamentally depends on the sensory errors and the delays in the system. GPS-INS (inertial navigation system) fusion could provide a method for more precise outdoor position and velocity estimation, while a control algorithm with more accurate feed-forward terms or learning capability could reduce the control relaxation time and thus the self-excited oscillations.

Our current algorithms are two and a half dimensional. Flocking and formation flights are performed in two dimensions but units vary their altitude during individual take-off and landing. Group members have the possibility to join or leave the flock any time in the third dimension and they do not have to interact if they have sufficient altitude gap between them. This behavior is similar to most natural aerial flocks, like V-shaped migrating birds, thermalling birds or paragliders. In theory, the flocking algorithm can be extended to three dimensions, but in practice the third, vertical dimension has to be treated quite differently in all cases. Due to gravity and the flight characteristics of quadcopters height control has completely different dynamics compared to position control.

With our current setup we could achieve a minimum of 6–10 m equilibrium distance between the robots in the 0–4 m/s velocity range. Considering ±2 m GPS accuracy, 1.5 s overall time lag and a random disturbance of wind, this is already the closest distance one can allow without the risk of collisions. In theory, the algorithms are not limited to the tested flock density or velocity range. With the dynamically tunable equilibrium distance a wide range of applications and flight situations can be handled. However, tighter flocks require better positioning accuracy, and higher speed implies a loosened flock to allow for sufficient 'breaking distance' between units.

The true advantage of a flocking flight over a single flying robot stands in its increased 'awareness', robustness and redundancy. The flock, as a meta-unit, can detect the environment more efficiently and can operate much longer than its members individually, such as cells in a living organism or migrating locusts over the ocean. The application potential of our system is large, ranging from ad-hoc mobile networks through distributed, self-organized monitoring of the environment (highway traffic, environmentally protected areas or agricultural lands) to stock delivery, rescue operation assistance, pest control, autonomous airport traffic control or even military applications. However, by demonstrating the stable flight of a truly autonomous, decentralized robotic flock, our main goal was to show that the various peaceful applications of drones are by now feasible.